\documentclass[letterpaper,10pt,conference]{ieeeconf}

\IEEEoverridecommandlockouts
\usepackage{balance}

\usepackage{amsmath,amssymb}
\usepackage{booktabs}
\usepackage{graphicx}
\usepackage{capt-of}
\usepackage{multirow}
\usepackage{tabularx}
\usepackage{array}
\usepackage{xcolor}
\usepackage{url}
\usepackage{microtype}
\usepackage{tikz}
\usetikzlibrary{arrows.meta,calc,fit,positioning}

\graphicspath{{figures/}}

\newcommand{\method}{StereoPatch}
\newcommand{\sptokens}{\emph{StereoPatch Tokens}}

\newcommand{\tabletitle}[1]{%
  \refstepcounter{table}%
  \par\begingroup\noindent\small\raggedright
  \textbf{TABLE~\Roman{table}:}~#1\par\endgroup\vspace{1.5mm}}
\newcommand{\figureplaceholder}[2]{%
  \setlength{\fboxsep}{0pt}%
  \fcolorbox{gray!65}{softgray}{%
    \parbox[c][#1][c]{0.985\linewidth}{%
      \centering\color{gray!75!black}\sffamily\footnotesize
      PLACEHOLDER\\[0.8mm]#2}}}
\newcolumntype{Y}{>{\raggedright\arraybackslash}X}

\definecolor{ink}{RGB}{37,41,37}
\definecolor{muted}{RGB}{98,104,98}
\definecolor{subtleborder}{RGB}{205,210,203}
\definecolor{rgbblue}{RGB}{111,134,191}
\definecolor{rgbpale}{RGB}{232,237,246}
\definecolor{depthorange}{RGB}{182,163,67}
\definecolor{depthpale}{RGB}{250,247,232}
\definecolor{fusiongreen}{RGB}{117,149,134}
\definecolor{fusionpale}{RGB}{228,237,232}
\definecolor{action}{RGB}{158,113,93}
\definecolor{actionpale}{RGB}{241,231,225}
\definecolor{softgray}{RGB}{243,243,240}

\DeclareMathSizes{7.3}{7.3}{5}{5}
\DeclareMathSizes{7.5}{7.5}{5}{5}
\DeclareMathSizes{7.6}{7.6}{5}{5}
\DeclareMathSizes{7.9}{7.9}{5}{5}

\title{\LARGE \bf
\method: Patch-Aligned RGB--Depth Fusion\\
for Spatial Perception in Robot Manipulation
}

\author{Yanan Zhou$^{1}$, Zhaoyan Qian$^{1}$, James Zhao$^{1}$, and Weiming Zhi$^{1,2,3,*}$%
\thanks{$^{1}$School of Computer Science and $^{2}$Australian Centre for Robotics, The University of Sydney, Australia.}%
\thanks{$^{*}$Corresponding author: \texttt{Weiming.Zhi@sydney.edu.au}.}}

\newcommand{\paperabstract}{
Recent advances in robot imitation learning have produced visuomotor policies that predict actions directly from visual observations. Yet visually similar scenes can require different actions as target position, object height, or contact geometry changes. Pretrained RGB features may map these geometrically distinct states to similar policy inputs, while simply adding depth requires the policy to learn RGB--depth correspondence from the same limited demonstrations used to learn control. We introduce \method{}, a patch-aligned RGB--depth representation that binds registered metric geometry directly to the RGB patches used for action prediction. On a shared 2-D patch grid, asymmetric cross-attention incorporates depth information into the corresponding RGB features before action decoding. The resulting \sptokens{} provide a geometry-aware visual representation that can condition general visuomotor policies without changing their underlying learning objectives. Across six real-robot tasks, \method{} achieves higher closed-loop success than appearance-only, geometry-only, raw RGB-D, and late-fusion baselines. Additional experiments across three simulation suites evaluate compatibility across visuomotor policy architectures, spatial generalization, and operating limits. Results suggest that resolving control-relevant geometric ambiguity benefits from aligning depth directly with the visual features used for action prediction, rather than supplying it as an independent modality. Project page: \url{https://aus.bot/research/stereopatch/}.
}

\begin{document}
\maketitle
\thispagestyle{empty}
\pagestyle{empty}

\suppressfloats[t]
\begin{abstract}
\paperabstract
\end{abstract}

\begin{figure}[t]
\centering
\includegraphics[width=\linewidth]{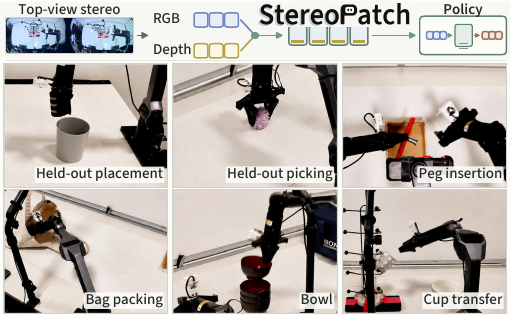}
\caption{\textbf{Spatial demands in robot learning.} Six real-robot tasks test position generalization, contact height, fine-scale and multi-stage control.}
\label{fig:teaser}
\end{figure}

\section{Introduction}

Imitation learning of visuomotor policies typically involves a visual encoder to transform camera observations into features for action prediction. Vision foundation models such as DINOv3~\cite{simeoni2025dinov3}, pretrained on large image collections, provide transferable RGB representations that reduce the need to learn visual features from limited robot demonstrations. However, recognizing an object does not determine how a robot should approach or contact it. Similar appearances can require different actions when target position, object height, or contact geometry changes. RGB features may underrepresent these metric differences even when they preserve object identity. This mismatch motivates improving spatial perception in the visual representations used by imitation policies.

Depth supplies complementary metric information, but its usefulness also depends on how it is represented. Recent work introduces DeFM~\cite{patel2026defm}, a foundation model pretrained entirely on depth images to learn transferable geometric and semantic features. This creates an opportunity to combine complementary RGB and depth features learned through large-scale pretraining. Yet providing two pretrained feature streams does not establish how they should interact. When supplied separately, the policy must still learn which RGB regions correspond to which depth regions from the same limited demonstrations used to learn control. The challenge is therefore to make their complementary information jointly useful for action prediction.

We introduce \method{}, a patch-aligned RGB--depth representation that combines these pretrained representations before action decoding. DINOv3 encodes RGB appearance, while DeFM encodes registered metric depth. Each RGB patch incorporates relevant depth features while retaining its image location, producing \sptokens{} that combine appearance and geometric evidence in one visual representation. This spatial association allows nearby geometry to inform a patch without imposing an exact one-to-one feature match. The tokens condition Action Chunking with Transformers (ACT)~\cite{zhao2023act} or Diffusion Policy (DP)~\cite{chi2023diffusion}. Both pretrained encoders remain frozen; the fusion module and action policy are trained jointly under the original imitation objective. Registered depth is obtained from calibrated stereo using Fast Foundation Stereo~\cite{wen2025fastfoundationstereo} or directly from an RGB-D camera.

\begin{figure*}[t]
\centering
\includegraphics[width=\textwidth]{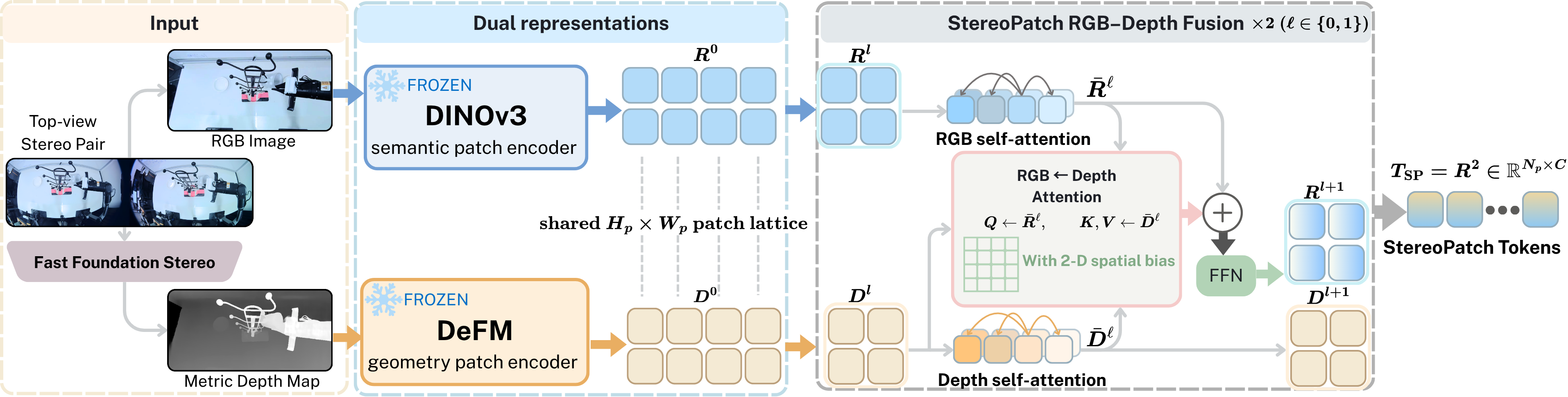}
\caption{\textbf{StereoPatch token construction.} Frozen DINOv3 and DeFM provide RGB and metric-depth features on a shared patch grid. Two asymmetric fusion blocks retrieve contextual geometry into RGB-indexed tokens, preserving their patch addresses for action prediction.}
\label{fig:token-construction}
\end{figure*}

We evaluate \method{} on six real-robot tasks and three simulation suites. Real-robot tasks test generalization to unseen locations, height-dependent contact, precise insertion, and changing task targets. Matched comparisons show higher observed closed-loop success than the tested modality and fusion controls. Simulation covers tabletop, multi-robot, and mobile manipulation with both action generators. Resolution sweeps, input swaps, and runtime measurements characterize operating limits. These results show that fusing pretrained appearance and metric geometry before action decoding improves closed-loop performance across these spatial manipulation tasks.

\noindent\textbf{The paper makes three contributions:}
{\setlength{\leftmargini}{1.45em}
\begin{itemize}
\setlength{\itemsep}{0.25ex}
\setlength{\parsep}{0pt}
\setlength{\topsep}{0.45ex}
\setlength{\partopsep}{0pt}
\item We identify a representation--control mismatch in visual imitation learning: similar RGB features can obscure geometric differences that require different robot actions.
\item We introduce \method{}, a policy-facing spatial interface that binds registered metric geometry to RGB patch addresses before action decoding, while remaining compatible with standard imitation objectives and action generators.
\item We evaluate \method{} across three simulation suites and six real-robot tasks covering spatial generalization, fine-scale contact, metric ambiguity, and multi-stage control. Matched real-robot comparisons record higher observed success than the tested modality and fusion controls.
\end{itemize}}

\section{Related Work}

\noindent\textbf{Visual representations and imitation learning:} ACT~\cite{zhao2023act} and Diffusion Policy~\cite{chi2023diffusion} provide expressive action generators, but their predictions depend on the spatial information retained in their visual inputs. Recent systems strengthen other parts of the imitation pipeline: TriManPolicy retimes demonstrations into synchronous tri-arm supervision, NestDex uses learned hand skills to assist dexterous demonstration collection, and PATCH monitors action-conditioned latent-patch innovation for deployment-time intervention~\cite{zhao2026trimanpolicy,zhao2026nestdex,zhou2026patch}. Representation methods instead transfer pretrained features, exploit action locality, or prioritize action-relevant tokens~\cite{nair2022r3m,simeoni2025dinov3,zhang2024sgrv2,zhang2026focusvla,liu2025vlapruner}. \method{} addresses a complementary question: it fuses RGB and depth features before action decoding, without changing demonstration collection or the policy's training objective.

\noindent\textbf{Geometric representations for manipulation:} Existing approaches represent geometry through PerAct's RGB-D voxels, RVT's virtual views, DP3/RISE point clouds, and SGR's back-projected RGB features~\cite{shridhar2022peract,goyal2023rvt,ze2024dp3,wang2024rise,zhang2023sgr}. Generalist policies inject ego-centric 3-D position encodings, point-cloud features, projected 3-D inputs, or a depth transformer~\cite{qu2025spatialvla,li2025pointvla,li2025bridgevla,yuan2025depthvla}; GAP and GeoPredict couple action learning to geometric evolution~\cite{xu2026gap,qian2026geopredict}. These families span explicit 3-D representations, large-scale policy pretraining, and auxiliary geometric prediction. \method{} studies how a compact imitation policy can gain metric sensitivity while retaining the image-patch organization of pretrained 2-D vision. Rather than replacing the RGB patch grid with voxels or point clouds, it enriches the RGB-indexed features with registered depth before action decoding.

\noindent\textbf{Stereo and RGB--depth token fusion:} StereoPolicy learns implicit correspondence from synchronized RGB views and studies stereo input, policy integration, camera configurations, and attention architectures~\cite{han2026stereopolicy}. EATR-Stereo preserves primary-view tokens and routes cross-view auxiliary tokens using proprioceptive history for humanoid VLA control~\cite{wu2026eatrstereo}. Our study is complementary: given registered metric depth and pretrained RGB/depth features, we examine how fusion affects learning across spatial control demands, using matched feature-level controls and task-specific diagnostics. \method{} retrieves DeFM's depth-native features into RGB-indexed tokens before action decoding~\cite{patel2026defm}. TokenFusion performs position-aligned token replacement, while DFormer learns RGB-D representations for semantic segmentation~\cite{wang2022tokenfusion,yin2024dformer}. We evaluate the resulting interface through closed-loop manipulation, using corresponding-patch MLP fusion as a direct comparison to \method{}'s fusion.

\section{StereoPatch: Patch-Aligned RGB--Depth Fusion}
\label{sec:method}

Figure~\ref{fig:token-construction} gives an overview of \method{}. StereoPatch takes registered RGB and metric depth and, before action decoding, produces geometry-aware tokens indexed by RGB patches. These tokens can condition different action policies together with wrist-view observations and robot state.

\subsection{Problem Formulation}

Given a robot demonstration dataset $\mathcal D$, each sample contains a reference RGB image $I$, its registered metric-depth map $Z$, and an expert action $A$. Let $E_R$ denote a pretrained RGB encoder. For some observation pairs $(i,j)$, it may hold that $E_R(I_i)\approx E_R(I_j)$ while $Z_i\not\approx Z_j$ and $A_i\not\approx A_j$: the RGB representations are similar although the geometric states and required actions differ. In this case, the RGB features supplied to the policy do not sufficiently expose the spatial distinctions relevant to control.

Simply adding depth does not directly resolve this ambiguity, because the policy must additionally learn correspondence between RGB and depth regions from the same limited demonstrations used to learn control. We therefore ask: how can metric geometry be aligned with the policy's RGB patch representation before action decoding?

At time $t$, in addition to $I_t$ and $Z_t$, the policy receives wrist-camera observations $W_t$ and robot state $s_t$, and predicts an action chunk $A_t=[a_t,\ldots,a_{t+H-1}]$ over horizon $H$. StereoPatch first constructs spatial tokens with a representation module $F_\theta$, after which an action policy $\pi_\phi$ predicts the action:
\begin{equation}
 T_t^{\mathrm{sp}}=F_\theta(I_t,Z_t), \qquad
 \widehat A_t=\pi_\phi(T_t^{\mathrm{sp}},W_t,s_t).
 \label{eq:policy-factorization}
\end{equation}
Here $T_t^{\mathrm{sp}}$ denotes StereoPatch tokens indexed by RGB patches. This factorization delimits the role of the method: StereoPatch changes the visual representation supplied to the policy, but does not prescribe how actions are generated.

\subsection{Aligned Appearance and Geometry Fields}

To resolve this ambiguity before action decoding, we first place appearance and geometry on a shared spatial coordinate system. A pretrained visual encoder converts an image into a 2-D grid of feature vectors. Each vector, or \emph{token}, summarizes a local image region; its row and column define its \emph{patch address}. We use separate encoders because RGB and depth carry different information. DINOv3 provides an appearance field organized by visual semantics, whereas DeFM converts metric depth into a geometry field~\cite{simeoni2025dinov3,patel2026defm}. The reference RGB frame $I_t$ comes from a calibrated top-view stereo rig, and Fast Foundation Stereo converts the synchronized pair into the registered metric-depth map $Z_t$~\cite{wen2025fastfoundationstereo}. An aligned RGB-D sensor could provide the same input interface.

Let $E_R$ and $E_D$ denote the frozen RGB and depth encoders, $\tau_R$ and $\tau_D$ their input transforms, and $\Pi_R$ and $\Pi_D$ trainable projections to a shared feature width $C$. We choose each branch's input size according to its native patch stride so that both encoders output the same $H_p\times W_p$ lattice:
\begin{equation}
 \begin{aligned}
 R_t^0&=\Pi_R E_R\!\left(\tau_R(I_t)\right), &
 D_t^0&=\Pi_D E_D\!\left(\tau_D(Z_t)\right),\\[-1pt]
 R_t^0,D_t^0&\in\mathbb R^{N_p\times C}, &
 N_p&=H_pW_p.
 \end{aligned}
 \label{eq:dual-fields}
\end{equation}
Only spatial patch tokens are retained; encoder-specific prefix tokens, which are global summaries without a unique image location, are removed. Because RGB and depth cover the same calibrated image domain, token index $p$ refers to the same normalized row--column cell in both fields. We call these fields \emph{co-indexed}: their addresses agree even though their feature values and the image context summarized by each encoder differ.

This difference makes direct addition inappropriate. A depth boundary may cross an RGB patch, and nearby depth context may be more informative than the feature at exactly the corresponding cell. Enforcing a one-to-one relation $R_{t,p}^0\leftrightarrow D_{t,p}^0$ would ignore differences in the encoders' receptive fields and feature organization. StereoPatch therefore uses soft retrieval over a shared patch-address system to select the geometric evidence relevant to each RGB patch.

\begin{table*}[!t]
\centering
\tabletitle{Simulation success over 100 in-house rollouts with distinct evaluation seeds; RoboMimic baselines are published percentages.}
\label{tab:simulation}
\scriptsize
\setlength{\tabcolsep}{2.6pt}
\renewcommand{\arraystretch}{0.90}
\begin{tabular*}{\textwidth}{@{\extracolsep{\fill}}llcccccccc@{}}
\toprule
& & \multicolumn{3}{c}{RoboMimic} & \multicolumn{3}{c}{RoboFactory} & \multicolumn{2}{c}{BEHAVIOR-1K} \\
\cmidrule(lr){3-5}\cmidrule(lr){6-8}\cmidrule(lr){9-10}
Perception & Method & ToolHang & Square & Transport & \shortstack{Lift\\Barrier} & \shortstack{Camera\\Align.} & \shortstack{3-Robot\\Stack} & \shortstack{Open\\Door} & \shortstack{Turn On\\Radio}\\
\midrule
\multirow{2}{*}{2-D images} & RGB-only DP & 53\% & 74\% & 92\% & 79/100 & 71/100 & 11/100 & 21/100 & 32/100\\
& Multi-view RGB DP & 54\% & 78\% & 92\% & 69/100 & 64/100 & 36/100 & 23/100 & 27/100\\
\midrule
\multirow{3}{*}{RGB-D / 3-D} & RGB-D & 56\% & 79\% & 94\% & 73/100 & 81/100 & 13/100 & 26/100 & 39/100\\
& RGBD-3DDA~\cite{ke2024diffuseractor} & 84\% & 83\% & 94\% & 65/100 & 77/100 & 9/100 & 39/100 & 33/100\\
& PCD-DP3~\cite{ze2024dp3} & 40\% & 69\% & 63\% & 52/100 & 79/100 & 31/100 & 27/100 & 29/100\\
\midrule
Stereo RGB & StereoPolicy-DP~\cite{han2026stereopolicy} & 94\% & 88\% & 94\% & 84/100 & 89/100 & 28/100 & 41/100 & 42/100\\
\midrule
\multirow{2}{*}{\textbf{Patch-aligned RGB-D}} & \textbf{\method{}-ACT} & 96/100 & 90/100 & 94/100 & 100/100 & 95/100 & 42/100 & 51/100 & 63/100\\
& \textbf{\method{}-DP} & 93/100 & 94/100 & 96/100 & 99/100 & 98/100 & 40/100 & 55/100 & 61/100\\
\bottomrule
\end{tabular*}
\par\smallskip
{\scriptsize\raggedright Demonstrations per task: 100 for our RoboMimic models and RoboFactory; 50 for BEHAVIOR-1K in OmniGibson.\par}
\end{table*}

\subsection{Retrieving Geometry into RGB Patches}

StereoPatch uses two fusion blocks ($\ell\in\{0,1\}$). Each block first applies self-attention independently within RGB and depth. In plain terms, self-attention lets a patch summarize other patches from the same modality before the two modalities interact. With a 2-D grid encoding $P$, these contextualized fields are
$\bar R_t^\ell=\operatorname{SA}_R^\ell(R_t^\ell;P)$ and
$\bar D_t^\ell=\operatorname{SA}_D^\ell(D_t^\ell;P)$; each operator includes normalization and a residual connection that retains its input.

The subsequent cross-attention retrieves relevant depth evidence for every RGB patch. For RGB index $p$, a \emph{query} describes what geometric evidence the appearance feature seeks. Each depth index $j$ provides a \emph{key}, used to score relevance, and a \emph{value}, the geometric feature transferred when selected. The matrices $W_Q$, $W_K$, and $W_V$ are learned linear projections, and $\operatorname{LN}$ denotes standard layer normalization. For one attention head,
\begin{equation}
 \begin{aligned}
 q_p^\ell&=W_Q^\ell\operatorname{LN}(\bar R_{t,p}^\ell+P_p),\\[-1pt]
 k_j^\ell&=W_K^\ell(\bar D_{t,j}^\ell+P_j),\quad
 v_j^\ell=W_V^\ell\bar D_{t,j}^\ell,\\[-1pt]
 \alpha_{pj}^\ell&=\operatorname{softmax}_{j}\!\left(
 \frac{q_p^\ell(k_j^\ell)^\top}{\sqrt d}+B_{pj}^\ell\right),\quad
 g_p^\ell=\sum_{j=1}^{N_p}\alpha_{pj}^\ell v_j^\ell,
 \end{aligned}
 \label{eq:geometry-retrieval}
\end{equation}
where $d$ is the per-head feature width. A larger relevance score gives the corresponding depth patch greater influence on the current RGB patch, and softmax normalizes the scores over all depth positions to weights that sum to one. Multi-head attention repeats this operation with different projections, allowing different heads to retrieve different geometric relations. We use $g_p^\ell$ below for their concatenated output.

Content similarity alone does not tell the model whether a depth feature lies at the same cell, one row above, or far across the image. We therefore add a learned 2-D relative-position term to every RGB--depth score:
\begin{equation}
 B_{pj}^{\ell}=b_{\mathrm{row}}^{\ell}[y_j-y_p]
 +b_{\mathrm{col}}^{\ell}[x_j-x_p].
 \label{eq:relative-position}
\end{equation}
Here $(x_p,y_p)$ and $(x_j,y_j)$ are the 2-D patch coordinates of the RGB and depth tokens. The two block-specific lookup tables encode row and column offsets, are shared across heads, and are learned through the imitation objective. This term is a spatial prior inside the forward pass, not an auxiliary loss.

Attention remains global over all $N_p$ depth tokens. The shared patch addresses make evidence from the same or neighboring locations identifiable, while global retrieval allows content to select more distant evidence when boundaries, occlusion, or contextual geometry make it useful. The construction therefore uses RGB--depth registration without imposing a fixed one-to-one feature match.

Cross-modal fusion is asymmetric: depth features update the RGB-indexed field, while the depth branch is updated through its own self-attention:
\begin{equation}
 \begin{aligned}
 \widetilde R_{t,p}^\ell&=\bar R_{t,p}^\ell+W_O^\ell g_p^\ell,\\[-1pt]
 R_{t,p}^{\ell+1}&=\widetilde R_{t,p}^\ell+
 \operatorname{FFN}^{\ell}(\operatorname{LN}(\widetilde R_{t,p}^\ell)),\\[-1pt]
 D_t^{\ell+1}&=\bar D_t^\ell.
 \end{aligned}
 \label{eq:fusion-update}
\end{equation}
Here $W_O^\ell$ maps the retrieved geometry back to the RGB feature space. The feed-forward network (FFN) is a small multilayer perceptron applied independently to each token, while the residual additions retain the original RGB information. After the second block, $T_t^{\mathrm{sp}}=R_t^2\in\mathbb R^{N_p\times C}$. Depth remains an internal evidence field rather than a second token sequence passed directly to the controller. Consequently, every output token retains its RGB patch address while carrying geometry selected by both feature content and relative location. In this paper, \emph{patch-aligned} denotes this shared address system and offset-aware retrieval; it does not mean hard one-to-one correspondence or a fixed patch resolution.

\subsection{Policy Integration}

\begin{figure}[!t]
\centering
\includegraphics[width=\columnwidth]{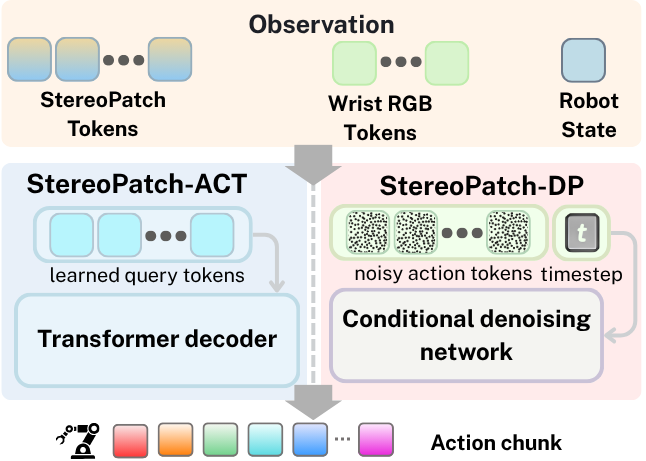}
\caption{\textbf{Policy integration.} StereoPatch tokens, wrist RGB, and robot state condition ACT or DP to predict action chunks under their original objectives.}
\label{fig:policy-instantiations}
\end{figure}

Figure~\ref{fig:policy-instantiations} shows how StereoPatch conditions an action generator. A wrist-view encoder provides local visual tokens $T_t^{\mathrm{w}}$, while $e_s(s_t)$ embeds joint and gripper state. We concatenate them with the StereoPatch tokens,
\begin{equation}
 O_t=[T_t^{\mathrm{sp}};T_t^{\mathrm{w}};e_s(s_t)],\qquad
 \widehat A_t=\pi_\phi(O_t),
 \label{eq:policy-observation}
\end{equation}
so $O_t$ is a unified observation sequence containing top-view appearance and geometry, wrist context, and robot state.

To test whether the representation transfers across action generators, we instantiate $\pi_\phi$ with Action Chunking with Transformers (ACT)~\cite{zhao2023act} and Diffusion Policy (DP)~\cite{chi2023diffusion}. ACT decodes the full chunk in parallel through learned action queries, one per predicted step. DP iteratively denoises an action chunk conditioned on $O_t$. Both retain their original action-generation structures.

Training retains each policy's native behavior-cloning objective $\mathcal L_{\pi}$, which matches the predicted action chunk to the demonstration:
\begin{equation}
 (\theta^*,\phi^*)=\arg\min_{\theta,\phi}
 \mathbb E_{\mathcal D}\!\left[\mathcal L_{\pi}(\widehat A_t,A_t)\right],
 \label{eq:training-objective}
\end{equation}
The pretrained RGB and depth encoders remain frozen; the input projections, fusion blocks, wrist-view and robot-state encoders, and action policy are optimized end-to-end. No RGB--depth correspondence labels or auxiliary fusion or positional losses are used. Thus, action supervision alone determines which geometric evidence enters each RGB patch, while ACT and DP retain their original objectives.

\section{Experiments}

We examine how spatial conditioning supports learning across positions (RQ2), metric contact (RQ3), and small targets and successive task stages (RQ4). Matched real-robot controls test the role of feature fusion; input swaps and response maps aid interpretation. Simulation evaluates the interface across action generators and manipulation settings (RQ1), and timing characterizes its inference budget (RQ5).

\noindent\textbf{Implementation and controls.} DINOv3 and DeFM remain frozen. By default, reference RGB inputs are $640\times320$ pixels (width$\times$height), with both feature fields on a $20\times40$ patch grid (rows$\times$columns). ACT uses a seven-layer decoder; DP retains its standard architecture. Our ACT and DP policies train for 80,000 steps with batch size 32, jointly optimizing fusion and policy modules. Fusion controls share encoders, grids, and action generators. \emph{Late concat.} passes RGB/depth features directly to the policy as separate tokens. \emph{Patch MLP} locally fuses corresponding features with a small MLP, $T_p=\operatorname{MLP}([R_p^0;D_p^0])$, without cross-modal attention.

\subsection{RQ1: Does StereoPatch Improve Performance Across Simulation Benchmarks?}

\begin{figure}[!ht]
\centering
\includegraphics[width=\columnwidth]{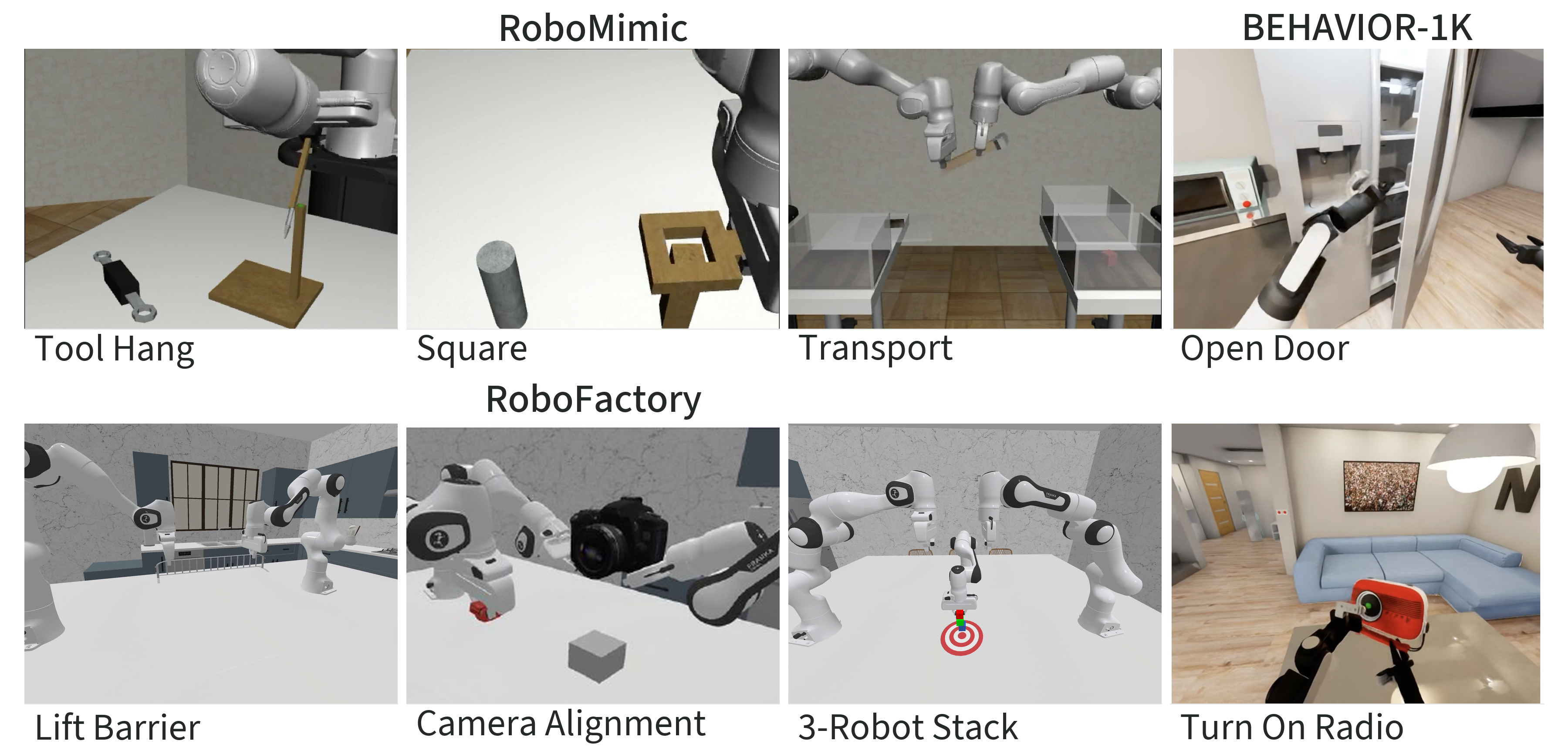}
\caption{\textbf{Simulation tasks.} RoboMimic, RoboFactory, and BEHAVIOR-1K cover tabletop, multi-robot, and mobile manipulation, respectively.}
\label{fig:simulation-tasks}
\end{figure}

Table~\ref{tab:simulation} reports results for the tasks in Figure~\ref{fig:simulation-tasks}. RoboMimic~\cite{mandlekar2021robomimic}, RoboFactory~\cite{qin2025robofactory}, and BEHAVIOR-1K~\cite{li2022behavior1k} test tabletop, multi-robot, and mobile manipulation, respectively. Using rendered RGB and depth, these settings test whether the interface remains useful beyond a single workspace or coordination structure.

\noindent\textbf{Baselines.} RGB-only DP uses monocular RGB from each view; Multi-view RGB DP concatenates features from both stereo images before Diffusion Policy~\cite{chi2023diffusion}. RGB-D adds depth. RGBD-3DDA (3D Diffuser Actor)~\cite{ke2024diffuseractor} uses a 3-D scene representation for action denoising; PCD-DP3~\cite{ze2024dp3} encodes sparse point clouds. StereoPolicy-DP~\cite{han2026stereopolicy} learns implicit correspondence from stereo RGB. Together, they compare image, explicit-geometry, and learned-stereo representations with our ACT and DP instantiations.

\textbf{The interface is useful with both action generators across the tested settings.} With DP fixed, \method{} exceeds StereoPolicy-DP on each evaluated RoboFactory and BEHAVIOR-1K task; on Turn On Radio, success rises from 42/100 to 61/100. ACT also performs well, with its relative performance against DP varying by task. These results support reusing the RGB--depth representation design across action generators and tasks. RoboMimic baseline rates are from StereoPolicy's original evaluation protocol~\cite{han2026stereopolicy}.

\subsection{RQ2: Does Patch-Aligned Geometry Improve Low-Data Spatial Generalization?}

\begin{figure}[!t]
\centering
\includegraphics[width=0.84\columnwidth]{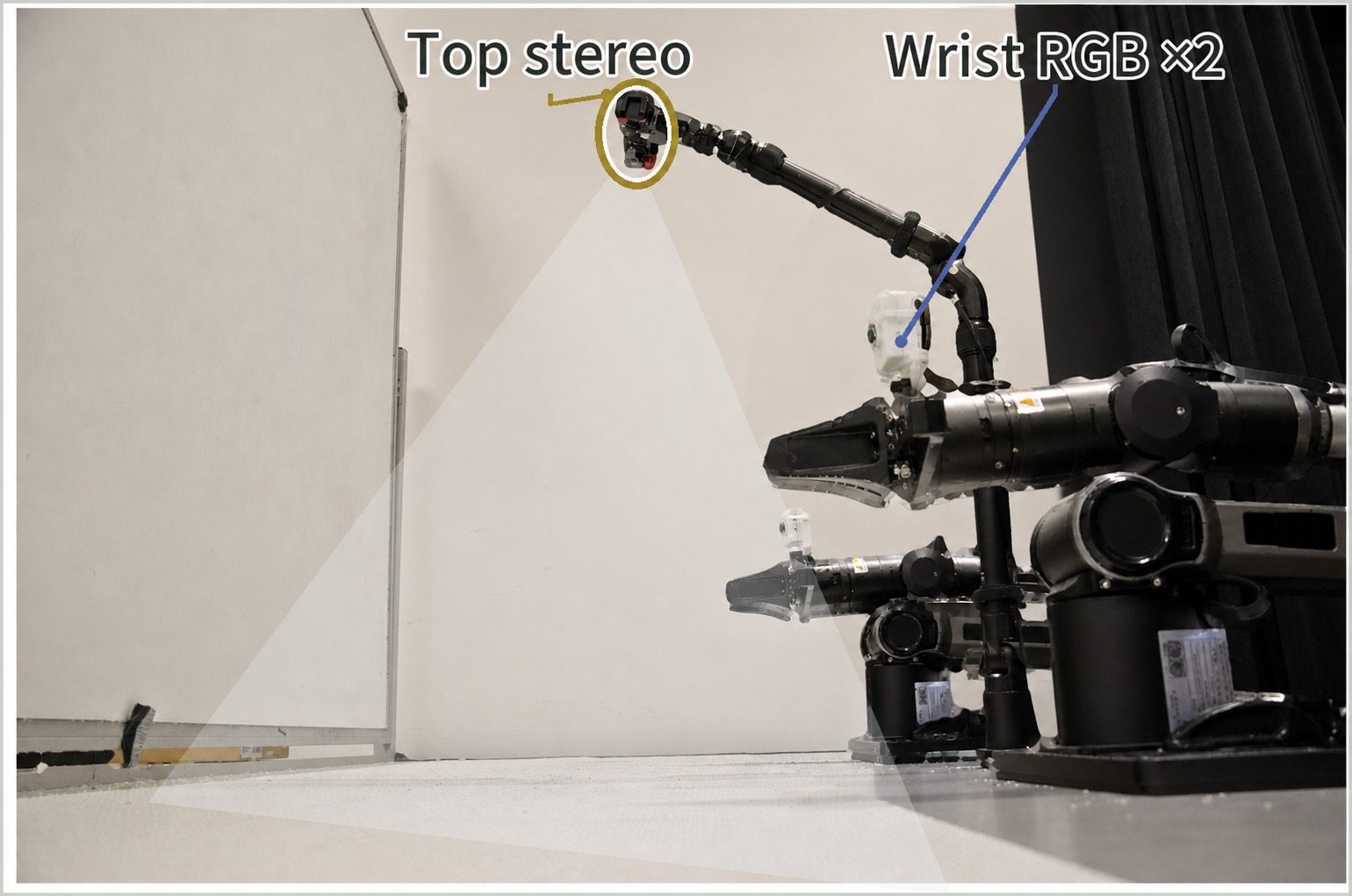}
\caption{\textbf{Robot sensing.} Top-view stereo provides RGB and depth; wrist RGB also conditions the policy. External views support scoring.}
\label{fig:real-robot-setup}
\end{figure}

Changing an object's position changes the required action while preserving the task and object identity. Placement and picking test this spatial adjustment from opposite sides: where to deliver an object and where to acquire it. Seen positions assess whether one policy can learn across demonstrated locations; held-out positions assess whether that mapping extends beyond the demonstrations.

\begin{figure}[t]
\centering
\includegraphics[width=\columnwidth]{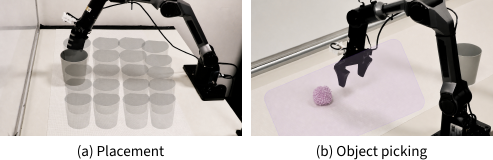}
\caption{\textbf{Spatial generalization tasks.} (a) Placement and (b) picking workspaces; held-out evaluations include extrapolation up to 5~cm.}
\label{fig:rq2-coverage}
\par\vspace{5pt}
\tabletitle{Placement and picking success at seen and held-out positions, with two closed-loop evaluation attempts per coordinate.}
\label{tab:rq2-position}
{\scriptsize
\setlength{\tabcolsep}{2.2pt}
\renewcommand{\arraystretch}{0.94}
\resizebox{\columnwidth}{!}{%
\begin{tabular}{@{}lccc@{}}
\toprule
\multicolumn{4}{c}{T1: Placement; $A=20$ demonstration anchors}\\
Method & Seen & Held-out & Combined\\
\midrule
$\pi_{0.5}$-FT (system)~\cite{physicalintelligence2025pi05} & 30/40 (75.0\%) & 28/40 (70.0\%) & 58/80 (72.5\%)\\
Appearance-only ACT & 4/40 (10.0\%) & 2/40 (5.0\%) & 6/80 (7.5\%)\\
Geometry-only ACT & 18/40 (45.0\%) & 12/40 (30.0\%) & 30/80 (37.5\%)\\
Patch MLP ACT & 29/40 (72.5\%) & 22/40 (55.0\%) & 51/80 (63.8\%)\\
\textbf{\method{}-ACT} & \textbf{34/40 (85.0\%)} & \textbf{32/40 (80.0\%)} & \textbf{66/80 (82.5\%)}\\
\midrule
\multicolumn{4}{c}{T2: Object picking; $A=37$ demonstration anchors}\\
Method & Seen & Held-out & Combined\\
\midrule
$\pi_{0.5}$-FT (system)~\cite{physicalintelligence2025pi05} & 62/74 (83.8\%) & 51/74 (68.9\%) & 113/148 (76.4\%)\\
Appearance-only DP & 19/74 (25.7\%) & 4/74 (5.4\%) & 23/148 (15.5\%)\\
Geometry-only DP & 34/74 (45.9\%) & 25/74 (33.8\%) & 59/148 (39.9\%)\\
Patch MLP DP & 49/74 (66.2\%) & 32/74 (43.2\%) & 81/148 (54.7\%)\\
\method{}-DP & 64/74 (86.5\%) & 59/74 (79.7\%) & 123/148 (83.1\%)\\
\textbf{\method{}-ACT} & \textbf{66/74 (89.2\%)} & \textbf{60/74 (81.1\%)} & \textbf{126/148 (85.1\%)}\\
\bottomrule
\end{tabular}}}
\end{figure}

The Piper setup uses a synchronized AR0234 global-shutter color stereo camera (Figure~\ref{fig:real-robot-setup}). Matched perception controls share demonstrations, action generators, evaluation states, resets, and success criteria. Each anchor supplies two training trajectories; each seen or held-out coordinate receives two evaluation attempts (Figure~\ref{fig:rq2-coverage}, Table~\ref{tab:rq2-position}). Our system reference, $\pi_{0.5}$-FT, fine-tunes $\pi_{0.5}$~\cite{physicalintelligence2025pi05} on these demonstrations.

\textbf{How the features interact matters especially at held-out positions.} In Table~\ref{tab:rq2-position}, Patch MLP improves on the single-modality controls, showing the value of associating corresponding appearance and geometry features. \method{} further improves held-out success by 25.0 percentage points for placement (ACT) and 36.5 for picking (DP). With encoders, grid, and action generator shared, depth availability and local correspondence alone do not recover this benefit. For both matched action generators, the gain over Patch MLP is larger at held-out than at seen positions. This pattern connects the benefit of RGB--depth fusion to spatial generalization beyond demonstrated locations. Low RGB-only success at seen positions also suggests difficulty learning position-dependent actions; spatial aliasing is a possible explanation~\cite{lee2026statealiasing}, as similar features may mask action differences.

\begin{figure}[t]
\centering
\IfFileExists{figures/rq2_data_efficiency.pdf}{%
  \includegraphics[width=\columnwidth]{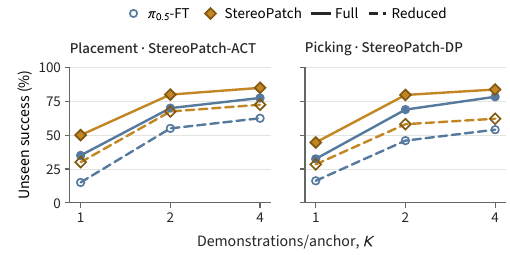}%
}{%
  \figureplaceholder{25mm}{Placement and object picking\\held-out success vs. trajectories per anchor $K$\\full and reduced training-anchor coverage}%
}
\caption{\textbf{Demonstration efficiency.} Held-out success versus trajectories per anchor, with varied training coverage and fixed evaluation positions.}
\label{fig:rq2-data-efficiency}
\end{figure}

Figure~\ref{fig:rq2-data-efficiency} examines how demonstration coverage and repetition affect this spatial learning. Compared with reduced coverage of 10 placement and 19 picking anchors, full coverage improves held-out success at the same number of trajectories per anchor. Increasing repetition yields diminishing gains, whereas broader coverage exposes more spatial conditions and also increases the total demonstration count. Across these settings, \method{} also exceeds $\pi_{0.5}$-FT~\cite{physicalintelligence2025pi05}, a system reference with different architecture and pretraining that complements the shared-feature fusion controls in Table~\ref{tab:rq2-position}.

\begin{table}[!htbp]
\centering
\tabletitle{Camera comparison on 74 seen picking trials, with separate data collection and StereoPatch-ACT training per camera.}
\label{tab:rq2-camera}
\scriptsize
\setlength{\tabcolsep}{5pt}
\renewcommand{\arraystretch}{0.94}
\begin{tabular*}{\columnwidth}{@{\extracolsep{\fill}}llr@{}}
\toprule
Camera & Depth source & Success $\uparrow$\\
\midrule
AR0234 stereo & Fast Foundation Stereo & 66/74 (89.2\%)\\
RealSense D435 & On-device depth & 54/74 (73.0\%)\\
RealSense D405 & On-device depth & 38/74 (51.4\%)\\
\bottomrule
\end{tabular*}
\end{table}

{\clubpenalty=10000
\textbf{Sensing setup affects closed-loop performance.} Table~\ref{tab:rq2-camera} evaluates the same \method{}-ACT architecture with three sensing setups. For each camera, we recollect demonstrations and retrain the policy, keeping the seen picking positions fixed. AR0234 stereo uses Fast Foundation Stereo to estimate depth; D435 and D405 supply device-generated depth. AR0234 achieves the highest success, followed by D435 and D405. Thus, both stereo-estimated and device-generated depth can condition the same policy architecture. The success differences make sensing configuration part of the practical design of the learned controller. Together, Tables~\ref{tab:rq2-position} and~\ref{tab:rq2-camera} examine two choices for spatial control: how RGB and depth features interact, and how the input observations are acquired.\par}

\subsection{RQ3: Does Patch-Aligned Geometry Help Under Metric Ambiguity?}

\begin{figure}[t]
\centering
\includegraphics[width=\columnwidth]{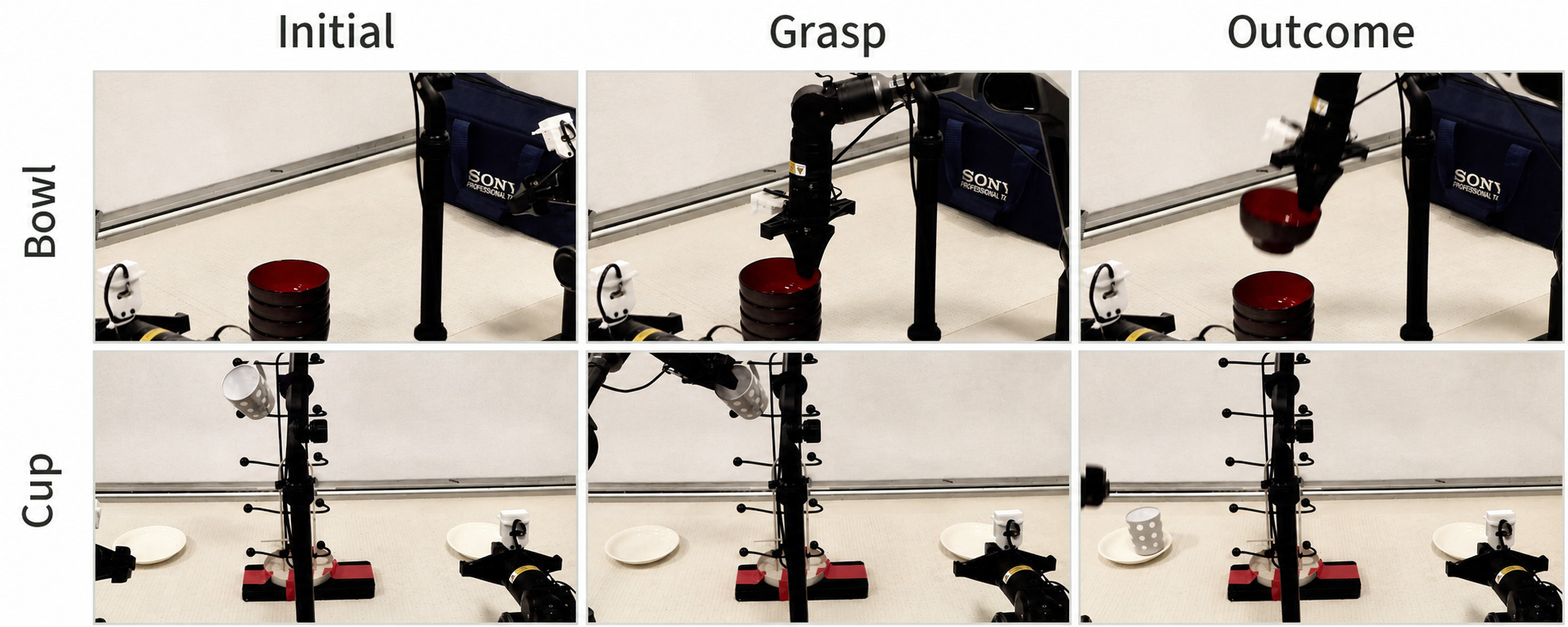}
\caption{\textbf{Metric contact tasks.} Bowl extraction varies contact height; cup transfer requires three-dimensional reaching across rack configurations.}
\label{fig:rq3-tasks}
\end{figure}

Knowing which object to grasp does not determine the required contact height. Bowl extraction varies stack height under similar top-view appearances, while cup transfer combines vertical and planar changes in the required reach (Figure~\ref{fig:rq3-tasks}). These tasks ask whether the spatial interface helps turn metric distinctions into appropriate contact, extending the position-learning study beyond changes in planar location.

\begin{figure}[t]
\centering
\includegraphics[width=\columnwidth]{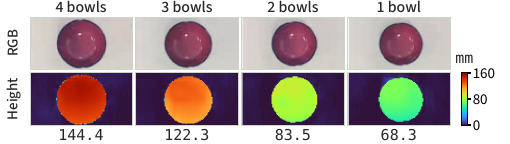}
\caption{\textbf{Height ambiguity.} Similar top-view RGB crops correspond to different bowl-stack heights, shown by metric height maps below (mm).}
\label{fig:rq3-visual-depth}
\end{figure}

The ACT controls distinguish depth availability from its use in the policy. Raw RGB-D and single-modality inputs test metric input and modality complementarity. Late concat. and Patch MLP share \method{}'s pretrained features, testing direct policy input and local fusion for metric contact.

Bowl trials cross four locations with stacks of one--four bowls; cup trials use four rack configurations. Matched ACT controls share demonstrations, evaluation conditions, checkpoint selection, and success criteria. Bowl success requires lifting exactly one bowl: grasping too high can miss the stack, while grasping too low can lift multiple bowls. These categories test whether higher success accompanies fewer errors above and below the useful contact region.

\begin{table}[!htbp]
\centering
\tabletitle{Bowl and cup success, with selected bowl failure categories: empty grasps (Empty) and multiple-bowl grasps (Multi.).}
\label{tab:rq3-depth}
\scriptsize
\setlength{\tabcolsep}{0.75pt}
\renewcommand{\arraystretch}{0.94}
\begin{tabular}{@{}>{\raggedright\arraybackslash}p{25.4mm}cccc@{}}
\toprule
& \multicolumn{3}{c}{Bowl extraction} & \multicolumn{1}{c}{Cup transfer}\\
\cmidrule(lr){2-4}\cmidrule(lr){5-5}
Method & Success $\uparrow$ & Empty $\downarrow$ & Multi. $\downarrow$ & Success $\uparrow$\\
\midrule
\multicolumn{5}{@{}l}{\emph{System references}}\\[-0.3mm]
$\pi_{0.5}$-FT & 65/80 (81.3\%) & 5/80 (6.3\%) & 10/80 (12.5\%) & 15/20 (75.0\%)\\
StereoPolicy-DP & 60/80 (75.0\%) & 7/80 (8.8\%) & 10/80 (12.5\%) & 16/20 (80.0\%)\\
\midrule
\multicolumn{5}{@{}l}{\emph{Matched ACT perception controls}}\\[-0.3mm]
Appearance-only & 46/80 (57.5\%) & 10/80 (12.5\%) & 20/80 (25.0\%) & 12/20 (60.0\%)\\
Raw RGB-D & 53/80 (66.3\%) & 8/80 (10.0\%) & 14/80 (17.5\%) & 14/20 (70.0\%)\\
Geometry-only & 56/80 (70.0\%) & 7/80 (8.8\%) & 12/80 (15.0\%) & 13/20 (65.0\%)\\
Late concat. & 59/80 (73.8\%) & 6/80 (7.5\%) & 10/80 (12.5\%) & 15/20 (75.0\%)\\
Patch MLP & 64/80 (80.0\%) & 5/80 (6.3\%) & 8/80 (10.0\%) & 16/20 (80.0\%)\\
\textbf{\method{}-ACT} & 69/80 (86.3\%) & 3/80 (3.8\%) & 5/80 (6.3\%) & \textbf{18/20 (90.0\%)}\\
\midrule
\multicolumn{5}{@{}l}{\emph{Cross-head variant}}\\[-0.3mm]
\textbf{\method{}-DP} & \textbf{78/80 (97.5\%)} & \textbf{0/80 (0.0\%)} & \textbf{2/80 (2.5\%)} & 17/20 (85.0\%)\\
\bottomrule
\end{tabular}
\end{table}

\begin{figure}[!t]
\centering
\includegraphics[width=\columnwidth]{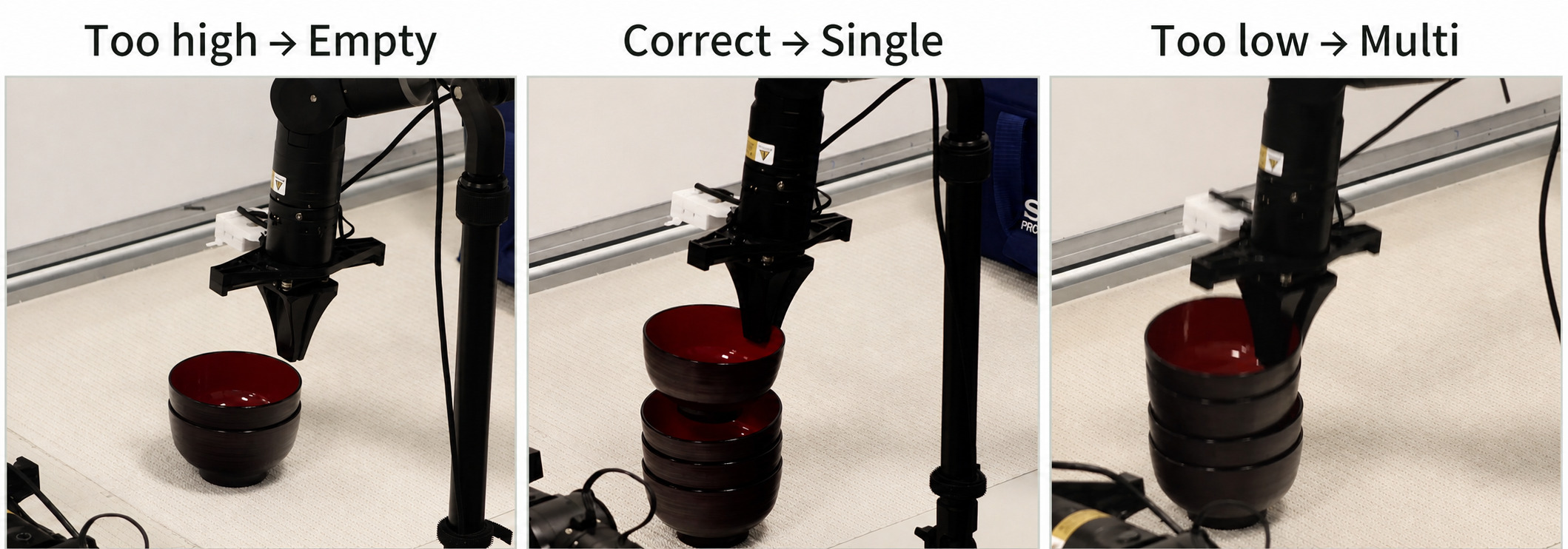}
\caption{\textbf{Bowl-contact outcomes.} Selected high, correct, and low grasps yield empty, single-bowl, and multiple-bowl pickup, respectively.}
\label{fig:rq3-signed-failures}
\end{figure}

\textbf{The fusion benefit extends to task-relevant contact outcomes.} In Table~\ref{tab:rq3-depth}, bowl success improves from late concatenation through Patch MLP to \method{}; the latter raises success from 64/80 to 69/80 over Patch MLP. Empty and multiple-bowl grasps both decrease (Figure~\ref{fig:rq3-signed-failures}), connecting the gain to fewer errors on either side of the useful contact region. The same success ordering is observed in cup transfer, where the required three-dimensional reach varies. With pretrained features and ACT fixed, these results show that \method{} helps turn differences in object height and 3-D position into effective contact and reaching actions.

Input swaps test sensitivity to height differences under similar RGB appearances (Figure~\ref{fig:rq3-visual-depth}). In three archived cases, robot state, wrist view, and background remain fixed. Depth swaps change the first ten predicted actions more than paired RGB swaps (2.9--5.5$\times$ in root-mean-square magnitude). Input swaps reveal action responses to depth changes; matched rollouts demonstrate gains in physical contact control.

\subsection{RQ4: How Does StereoPatch Behave Under Fine-Scale and Multi-Stage Control?}

Small targets and successive objects challenge spatial conditioning. Peg insertion tests acquisition and precise contact; picnic-bag packing tests progress across objects. Cumulative milestones relate final success to earlier progress.

\begin{figure}[!tbp]
\centering
{\footnotesize\sffamily\bfseries\color{ink}(a) Fine-scale peg insertion\par}
\includegraphics[width=\columnwidth]{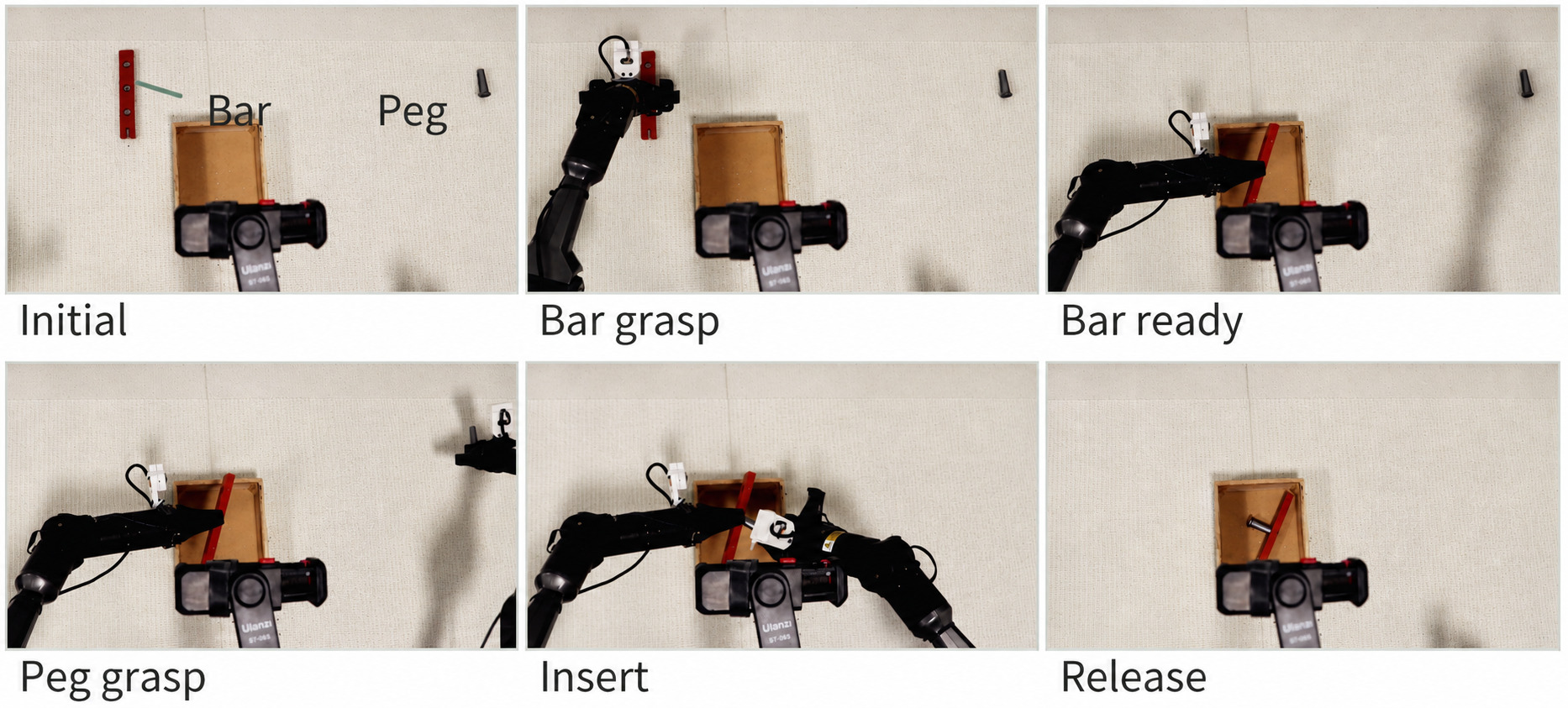}
\par\smallskip
{\footnotesize\sffamily\bfseries\color{ink}(b) Token-grid support\par}
\includegraphics[width=\columnwidth]{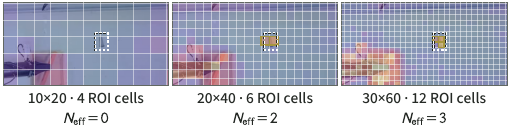}
\caption{\textbf{Fine-scale insertion.} (a) Task sequence. (b) Small-target support across three patch grids; maps are independently normalized.}
\label{fig:rq4-peg-evidence}
\end{figure}

\paragraph{Fine-scale peg insertion}
A finer grid increases a small target's spatial support, but the policy must learn to use it. We vary the grid with DP fixed (Figure~\ref{fig:rq4-peg-evidence}(a)). Each condition crosses six bar and six peg locations with two attempts per pair. Pick Peg requires prior bar pickup; Task adds insertion, revealing where progress is lost.

\begin{table}[t]
\centering
\tabletitle{Peg-insertion success across patch grids with DP: cumulative grasp and task success over 72 trials per condition.}
\label{tab:rq4-peg}
\scriptsize
\setlength{\tabcolsep}{2.0pt}
\renewcommand{\arraystretch}{0.92}
\begin{tabular}{llrrr}
\toprule
Perception & Grid & Pick Bar $\uparrow$ & Pick Peg $\uparrow$ & Task $\uparrow$\\
\midrule
RGB-only & $10\!\times\!20$ & 13/72 (18.1\%) & 4/72 (5.6\%) & 1/72 (1.4\%)\\
RGB-only & $20\!\times\!40$ & 17/72 (23.6\%) & 6/72 (8.3\%) & 3/72 (4.2\%)\\
RGB-only & $30\!\times\!60$ & 15/72 (20.8\%) & 5/72 (6.9\%) & 2/72 (2.8\%)\\
Geometry-only & $20\!\times\!40$ & 27/72 (37.5\%) & 22/72 (30.6\%) & 15/72 (20.8\%)\\
\method{}-DP & $10\!\times\!20$ & 25/72 (34.7\%) & 19/72 (26.4\%) & 13/72 (18.1\%)\\
\textbf{\method{}-DP} & $\mathbf{20\!\times\!40}$ & \textbf{37/72 (51.4\%)} & \textbf{32/72 (44.4\%)} & \textbf{26/72 (36.1\%)}\\
\method{}-DP & $30\!\times\!60$ & 30/72 (41.7\%) & 24/72 (33.3\%) & 18/72 (25.0\%)\\
\bottomrule
\end{tabular}
\end{table}

\textbf{The combined representation improves progress before final insertion.} Table~\ref{tab:rq4-peg} shows that, at the intermediate grid, \method{} improves both grasp milestones and completion over the single-modality controls. Its advantage over RGB-only persists across the tested resolutions. The representation improves object acquisition and task completion, linking final success to progress through the sequence.

\textbf{Finer spatial sampling does not guarantee better control.} Success peaks at the intermediate grid, then falls from 36.1\% to 25.0\% at the finest grid despite greater target coverage. A finer grid exposes more detail while changing the token sequence that action supervision must organize. The intermediate grid achieves the best closed-loop performance among the tested resolutions, motivating task-dependent selection of spatial sampling.

Figure~\ref{fig:rq4-peg-evidence}(b) illustrates target support. For an $H\times W$ image and projected target region $\Omega_t$, the nominal footprint is $\rho_t=|\Omega_t|H_pW_p/(HW)$ cells; $N_{\mathrm{eff}}$ counts cells above a local-background threshold. Independently normalized maps show target support across grids.

\begin{figure}[!tbp]
\centering
{\footnotesize\sffamily\bfseries\color{ink}(a) Two-stage picnic-bag control\par}
\includegraphics[width=\columnwidth]{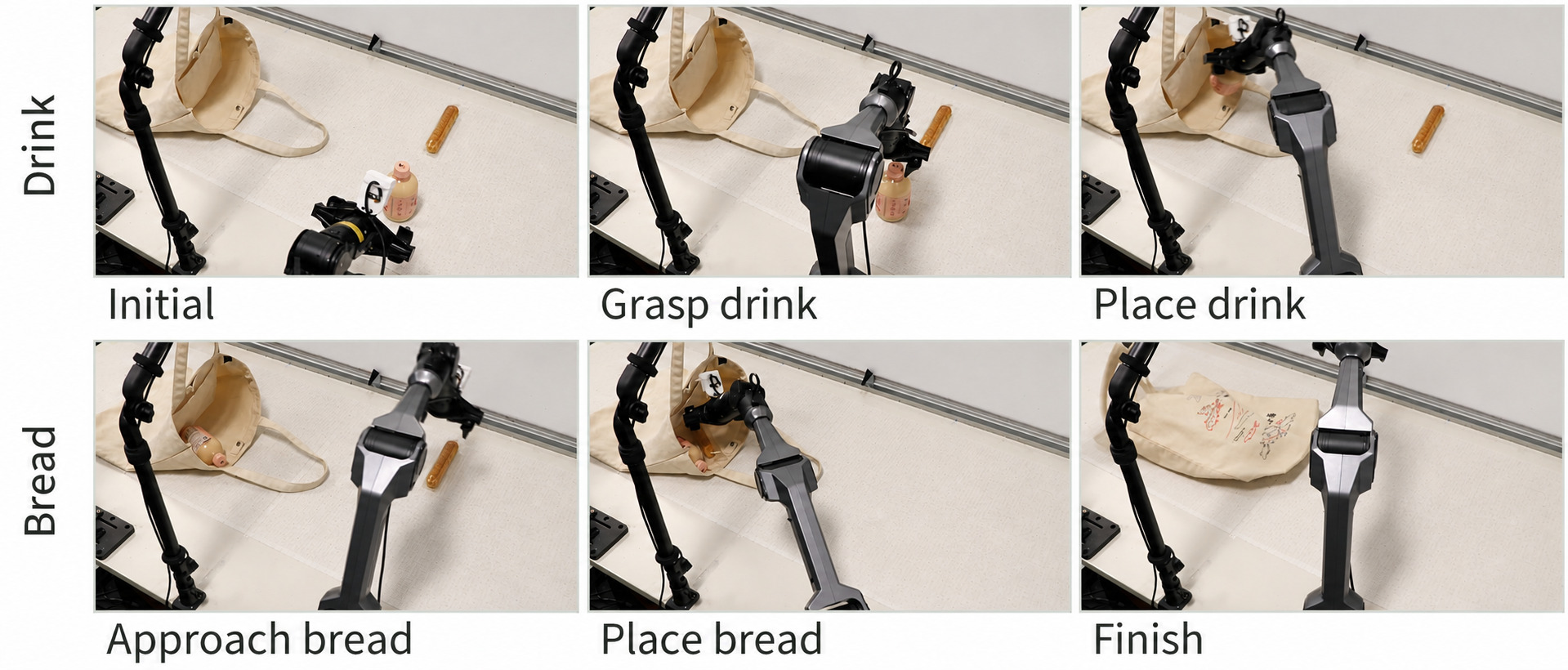}
\par\smallskip
{\footnotesize\sffamily\bfseries\color{ink}(b) Stage-conditioned action-response maps\par}
\includegraphics[width=\columnwidth]{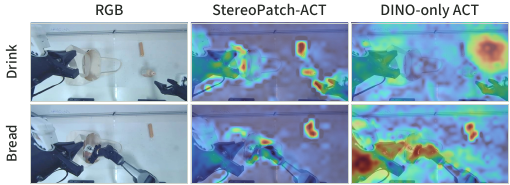}
\caption{\textbf{Multi-stage packing.} (a) Drink-then-bread sequence. (b) Selected action-response maps; each map is independently normalized.}
\label{fig:rq4-picnic-evidence}
\end{figure}

\paragraph{Multi-stage control}
The picnic-bag task requires placing a drink before selecting and manipulating bread (Figure~\ref{fig:rq4-picnic-evidence}(a)). First-object success can obscure difficulty with later subtasks. Table~\ref{tab:rq4-picnic} uses a common denominator for initial acquisition, later progress, and task completion.

\begin{table}[!htbp]
\centering
\tabletitle{Picnic-bag progress with ACT: cumulative drink-grasp, bread-grasp, and task success over 24 trials per method.}
\label{tab:rq4-picnic}
\scriptsize
\setlength{\tabcolsep}{2.0pt}
\renewcommand{\arraystretch}{0.92}
\begin{tabular}{@{}>{\raggedright\arraybackslash}p{35mm}ccc@{}}
\toprule
Perception & Drink grasp $\uparrow$ & Bread grasp $\uparrow$ & Task success $\uparrow$\\
\midrule
Appearance-only ACT & 10/24 (41.7\%) & 3/24 (12.5\%) & 1/24 (4.2\%)\\
Geometry-only ACT & 8/24 (33.3\%) & 5/24 (20.8\%) & 2/24 (8.3\%)\\
Late concat. ACT & 11/24 (45.8\%) & 4/24 (16.7\%) & 2/24 (8.3\%)\\
Patch MLP ACT & 13/24 (54.2\%) & 6/24 (25.0\%) & 4/24 (16.7\%)\\
\textbf{\method{}-ACT} & \textbf{15/24 (62.5\%)} & \textbf{8/24 (33.3\%)} & \textbf{7/24 (29.2\%)}\\
\bottomrule
\end{tabular}
\end{table}

\textbf{Fusion supports progress beyond the first object.} In Table~\ref{tab:rq4-picnic}, \method{} achieves the highest cumulative bread-grasp and completion counts among the matched ACT controls, completing 7/24 trials. The drop between drink and bread grasp reflects the difficulty of transport, placement, and acquiring the next object. Its advantage in subsequent grasping and final completion shows that the fusion benefit extends beyond the first object.

The selected action-response maps concentrate on the drink and bread at their respective grasp decisions (Figure~\ref{fig:rq4-picnic-evidence}(b)). The score $s_{t,i}=\|u_{t,i}\odot\nabla_{u_{t,i}}(\tfrac12\|\widehat A_t\|_F^2)\|_1$ uses token features $u_{t,i}$ to measure feature-weighted sensitivity of predicted action magnitude. At these two grasp decisions, the independently normalized maps illustrate visual information use associated with the current object.

\par\addvspace{2pt}
\noindent\begin{minipage}{\columnwidth}
\centering
\setlength{\abovecaptionskip}{1pt}
\includegraphics[width=0.80\columnwidth,trim=0 2pt 0 3pt,clip]{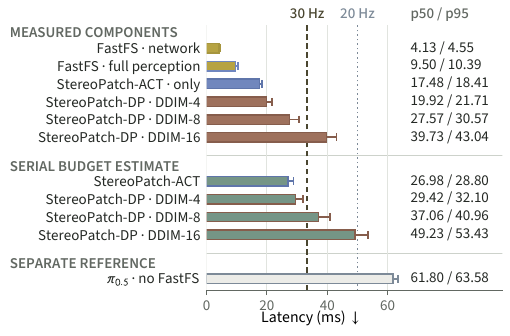}
\captionof{figure}{\textbf{Inference time.} Component p50/p95 measurements and their serial sums on an RTX~5090; sums estimate the inference budget.}
\label{fig:rq5-runtime}
\end{minipage}
\par\nopagebreak[4]
\subsection{RQ5: What Is the Inference-Time Operating Budget?}

The representation's cost must be considered together with the action generator. Figure~\ref{fig:rq5-runtime} profiles warm batch-1 inference on an otherwise idle RTX~5090. \textbf{ACT and four-step DP accommodate perception within the displayed 30-Hz budget.} Longer denoising schedules consume this margin, making decoder configuration part of the operating choice. The p50/p95 values denote median and 95th-percentile latency. Serial budgets are estimated by summing component p50/p95 values to compare the inference costs of perception and action-generation configurations. Deployment must also budget for resource contention, queueing, data transfer, and observation age; the separate $\pi_{0.5}$ reference excludes FastFS.

\section{Conclusion}

\method{} studies spatial perception through the actions a robot must learn: generalizing reaches, adjusting metric contact, manipulating small targets, and progressing between task targets. Its RGB-indexed fusion interface improves held-out success over corresponding-patch MLP fusion by 25.0 and 36.5 percentage points with encoders, grid, and action generator held fixed. Contact and stage-wise results extend this evidence, while non-monotonic resolution gains show that finer sampling alone is insufficient. The practical implication is to treat the interaction between pretrained appearance and geometry as part of policy design. \textbf{Future work.} We will focus on improving fine-scale and multi-stage task completion, exploring stronger RGB and depth encoders, and addressing coupled viewpoint, navigation, and workspace variation in mobile manipulation.

\balance
\bibliographystyle{IEEEtran}
\bibliography{references}

\end{document}